\documentclass[journal]{IEEEtran}
\usepackage{dsfont}

\usepackage[utf8]{inputenc}
\usepackage{cite}
\usepackage{amsmath,amssymb,bm}
\usepackage{graphicx}
\usepackage{xcolor}
\usepackage{booktabs,multirow,array}
\usepackage{url}
\usepackage[colorlinks=true,citecolor=blue,urlcolor=black]{hyperref}

\graphicspath{{figures/}}


\newcommand{\gA}[1]{\addlinespace[0.5ex]\multicolumn{12}{@{}l}{\textit{\textcolor{black!65}{#1}}}\\[0.3ex]}

\newcommand{\M}[1]{\hspace{0.9em}#1}
\newcommand{\R}{\mathbb{R}}

\newcommand{\OmegaSet}{\Omega}

\newcommand{\logits}{\boldsymbol{z}}
\newcommand{\temp}{\tau}

\newcommand{\ECE}{\operatorname{ECE}}
\newcommand{\SCE}{\operatorname{SCE}}
\newcommand{\ACE}{\operatorname{ACE}}

\newif\ifdraftnotes
\draftnotesfalse

\begin{document}
\bstctlcite{IEEEexample:BSTcontrol}

\title{CARD: Calibration via Agreement in Reverse Diffusion for Out-of-Domain MRI Segmentation}

\author{Jiaheng Dai, Weidong Guo, Qingbiao Li, Jie Xu, Yi Guo, Yuanyuan Wang and Zeju Li%
\thanks{Jiaheng Dai, Weidong Guo, Jie Xu, Yi Guo, Zeju Li, and Yuanyuan Wang are with the College of Biomedical Engineering, Fudan University, Shanghai, China. Qingbiao Li is with the Faculty of Information Science and Computing, University of Macau, Macau, China.}%
\thanks{Corresponding authors: Yi Guo, Yuanyuan Wang and Zeju Li (e-mail: zejuli@fudan.edu.cn).}%
}

\markboth{IEEE Transactions on Medical Imaging}%
{Dai \MakeLowercase{\textit{et al.}}: CARD: Calibration via Agreement in Reverse Diffusion}

\maketitle

\begin{abstract}
Probability calibration aligns model confidence with predictive accuracy, enabling clinicians to identify unreliable segmentation regions.
This alignment breaks down under domain shift, where artifacts and unseen protocols produce confident errors.
Existing post-hoc methods adapt the correction at test time, conditioning on predictive entropy, the logit pattern, or augmentation response, but each proxy is read from the terminal prediction, the very quantity that shift corrupts.
This motivates reliability evidence beyond the terminal prediction, which categorical diffusion provides in two ways. First, a generative shape prior keeps a capacity-limited reference intact when appearance is corrupted, so its disagreement with the primary segmentor highlights primary-model errors. Second, every reverse step yields a class distribution, separating persistent disagreement from
transient discrepancy.
Aggregated over the trajectory, this disagreement correlates with Dice at 0.788, against 0.521 for a matched discriminative control.
We therefore propose CARD (\textbf{C}alibration via \textbf{A}greement in \textbf{R}everse \textbf{D}iffusion), which maps the temporal aggregate of this disagreement to a temperature field applied per pixel across all classes, so that confidence changes while the segmentation does not. 
Across cardiac, prostate and brain MRI shifts, CARD lowers calibration error in 45 of 49 comparisons against the strongest baseline in each setting.
\footnote{Code is available at \url{https://github.com/fdu-farm/CARD}.}

\end{abstract}


\IEEEpeerreviewmaketitle

\section{Introduction}
\label{sec:introduction}

Modern segmentation networks are accurate, yet often produce miscalibrated and overconfident probability estimates.
Probability calibration addresses this mismatch by aligning predicted probabilities with empirical pixel-wise accuracy~\cite{guo2017calibration,mehrtash2020confidence}.
Miscalibration becomes more pronounced under out-of-domain (OOD) shifts in magnetic resonance imaging (MRI), where imaging artifacts and unseen acquisition protocols often lead to localized errors with high confidence.

\begin{figure*}[!t]
\centering
\includegraphics[width=\textwidth]{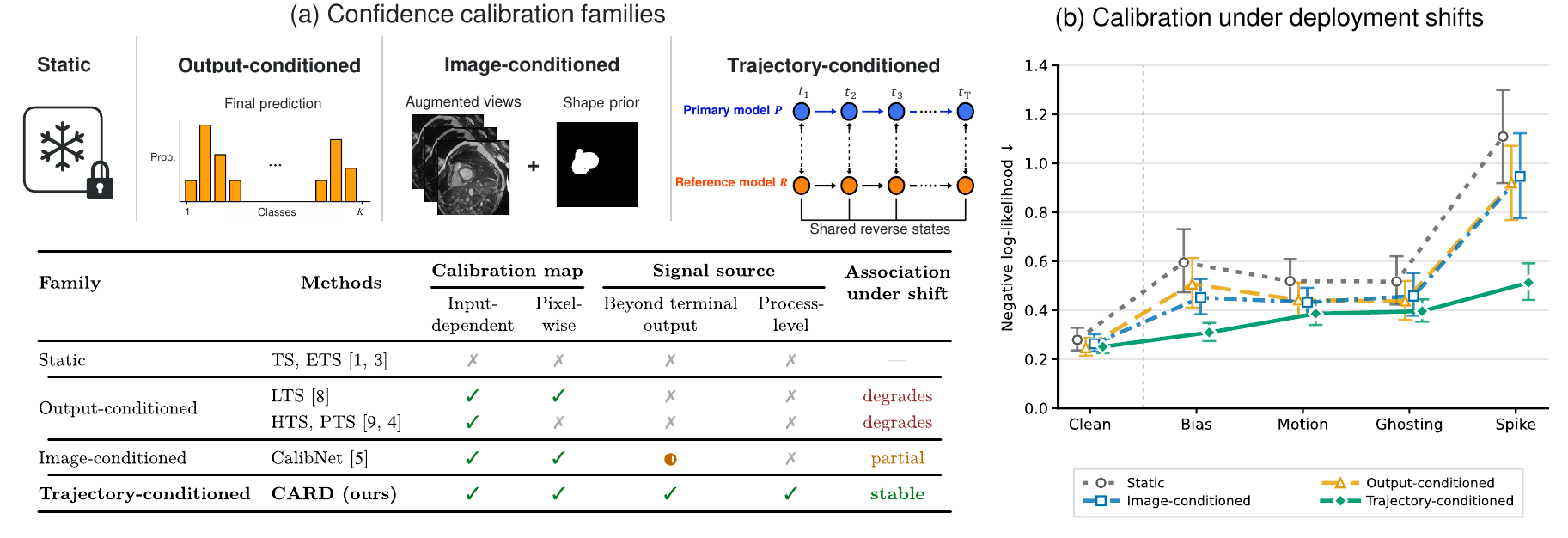}
\caption{
\textbf{CARD derives reliability evidence from the prediction process.}
\textbf{(a)} Existing families take their evidence from the final prediction, or from nothing at all: static methods apply a frozen correction, output-conditioned methods read the terminal output, and image-conditioned methods add augmented views and a shape prior. CARD instead compares a primary and a capacity-limited reference across shared reverse states, the only evidence source in the table that is not terminal.
\textbf{(b)} NLL on clean cardiac MRI and four artifact shifts. CARD stays lowest across every corruption, and the margin widens with severity.
}
\label{fig:calibration_signals}
\end{figure*}

Probability calibration for medical image segmentation faces two practical challenges.
First, calibration errors are spatially heterogeneous, with high-confidence errors often concentrated near anatomical boundaries and small or sparse foreground regions.
A global temperature applies the same correction across all pixels and cannot account for this spatial variation.
Second, both the location and severity of miscalibration vary across test cases, so the correction has to adapt to the current prediction instead of remaining fixed~\cite{mehrtash2020confidence,gal2016dropout,wang2019aleatoric}.

Existing post-hoc calibration methods differ in what they condition on at deployment.
\emph{Static calibration}, including temperature scaling (TS) and ensemble temperature scaling (ETS), applies a fixed correction fitted on labeled source-validation data~\cite{guo2017calibration,zhang2020mix}.
\emph{Output-conditioned calibration}, including local temperature scaling (LTS), Entropy-based Temperature Scaling (HTS), and Parameterized Temperature Scaling (PTS), makes the correction depend on the terminal model output~\cite{ding2021local,balanya2024adaptive,tomani2022parameterized}. 
\emph{Image-conditioned calibration}, represented by CalibNet, adds evidence extracted from the test image itself~\cite{ouyang2022improved}. Section~\ref{sec:related} describes each family in detail, and Fig.~\ref{fig:calibration_signals}(a) compares them by the evidence available at deployment and the freedom of the resulting correction.

Across these families the evidence is read from the terminal prediction, and its relationship to error is established on source data.
Under shift, confidence, entropy, logits, and augmentation responses change without preserving that relationship, and the mismatch varies across test cases. 
Calibration therefore needs evidence that requires no test labels, varies across cases and locations, and stays informative as appearance changes, which points beyond the endpoint to the prediction process.

Diffusion-based segmentation offers a natural source of such evidence, as predictions are formed through a sequence of intermediate reverse states~\cite{ho2020denoising,wu2024medsegdiff}. 
We find that capacity-limited Diffusion Probabilistic Models (DPMs) preserve consistent shape representations across varying domains and acquisition conditions, providing a structurally stable comparator when image appearance changes under distribution shift.

Prior diffusion-based work characterizes ambiguity or uncertainty through multiple sampled predictions~\cite{ahn2025modiff,wang2025caldiff}. We instead reuse a single reverse denoising trajectory.
CARD pairs a high-capacity primary diffusion model with a capacity-limited reference and measures their disagreement across shared intermediate reverse states.
The aggregated agreement captures how prediction contrast evolves during reverse refinement and remains associated with segmentation error across the evaluated shifts. Fig.~\ref{fig:calibration_signals}(b) shows the resulting calibration behavior under controlled cardiac artifact shifts.



Our contributions can be summarized as follows:
\begin{itemize}
    \item We present CARD, a post-hoc test-time calibration framework for diffusion-based segmentation that converts primary and reference disagreement across shared reverse states into a pixel-wise, input-dependent temperature field without updating the segmentor.

    \item We identify trajectory disagreement between a primary and a capacity-limited reference as a test-time reliability signal, and show that its association with segmentation error remains stable across the evaluated domain shifts, where terminal-output evidence does not.

   \item We introduce Temporal Aggregation Calibration (TAC), which aggregates discrepancies observed at multiple reverse states into a single spatial calibration map.
   
    \item We evaluate CARD on cardiac artifact and cross-vendor shifts, prostate source-to-target transfer, and brain-lesion scanner shift, lowering calibration error in 45 of 49 comparisons against the strongest baseline in each setting.
\end{itemize}

\section{Related Work}
\label{sec:related}

\subsection{Calibration for Medical Segmentation}

Training-time calibration modifies the segmentation objective or data distribution during model optimization.
The choice of objective can directly affect probability reliability; for example, Dice-based training may improve segmentation overlap while degrading calibration relative to cross-entropy~\cite{mehrtash2020confidence}.
Prior work has explored focal-loss variants, label smoothing, mixup, adversarial calibration losses, segmentation-specific objectives~\cite{mukhoti2020calibrating,szegedy2016rethinking,thulasidasan2020,tomani2021towards,karimi2023tait_calib_ood_seg}.
Recent differentiable calibration losses formulate marginal calibration error as a per-image auxiliary objective, allowing probability reliability to be optimized jointly with segmentation accuracy~\cite{barfoot2026average}.

Post-hoc calibration adjusts the probability outputs of a fixed segmentor after training.
Earlier methods map classifier scores to calibrated probabilities using learned transformations~\cite{platt1999probabilistic,zadrozny2001obtaining}.
TS learns a single scalar temperature from held-out data and applies it uniformly to the logits, preserving the predicted class ordering~\cite{guo2017calibration}.
ETS combines multiple calibrated components to increase the flexibility of the correction while retaining the original prediction~\cite{zhang2020mix}.
Both fit their parameters once on source-validation data and apply them unchanged at deployment.

Spatial calibration relaxes the assumption that one correction suits every location.
Mask-guided temperature scaling concentrates calibration learning on predicted foreground and candidate lesion regions, reducing the contribution of correctly classified background pixels to the fitting objective~\cite{zhang2024maskts}.
Spatial calibrators learned on source-validation data may nonetheless fail to transfer across changes in scanner, acquisition protocol, or image appearance~\cite{ovadia2019canyoutrust,gong2021confidence}.

\subsection{Test-time Adaptive Calibration}

Output-conditioned calibration makes the correction depend on the model output available at deployment.
LTS uses a convolutional calibrator that receives the image and segmentation logits and predicts a temperature field over the image, giving a spatially varying correction that still reads only the terminal output~\cite{ding2021local}.
HTS maps normalized predictive entropy to a prediction-specific temperature~\cite{balanya2024adaptive}, whereas PTS retains the full pattern of value-sorted logits and predicts the temperature through a compact neural network~\cite{tomani2022parameterized}.
HTS therefore summarizes the terminal output through a scalar, while PTS uses the relative magnitudes of the complete logit vector.
In dense segmentation, the corresponding mapping can be evaluated for each pixel-wise class distribution.

Image-conditioned calibration introduces additional evidence from the current image.
Ouyang et al.~\cite{ouyang2022improved} estimate pixel-level susceptibility from the mean and variance of logits obtained under repeated photometric augmentations.
A denoising autoencoder maps the uncalibrated prediction to a plausible anatomical shape, and the residual from the original prediction identifies deviations from the learned shape prior.
A calibration network receives the image, logits, perturbation statistics, and shape residual, and estimates a pixel-wise temperature map.
Both the augmentation response and the shape prior are established on the source domain, so their relation to segmentation error is not guaranteed to hold once acquisition conditions change.

\subsection{Reliability Estimation for Segmentation}

Uncertainty estimation and probability calibration address related questions.
Stochastic forward passes and model ensembles measure variability across predictions, while probabilistic segmentation models and variance-prediction networks represent ambiguity or data-dependent noise~\cite{gal2016dropout,lakshminarayanan2017ensembles,baumgartner2019phiseg,kendall2017uncertainties,wang2019aleatoric}.
Probability calibration instead evaluates whether predicted class probabilities agree with empirical accuracy.

Reliability cues have been used in medical image segmentation for quality control, case triage, and failure detection~\cite{robinson2017automatic,devries2018leveraging,roy2019bayesian,mehrtash2020confidence,li2022towards,wang2020deep}.
Image-level quality estimators predict case- or structure-level scores, while pixel-level estimators locate likely segmentation errors.
Li et al.~\cite{li2022towards} construct a self-reflective reference by reconstructing the input image from the predicted segmentation, and use
the discrepancy between the reconstruction and the original image to estimate per-class segmentation quality and a pixel-wise correctness map.
Such auxiliary-reference and self-evaluation approaches produce quality scores or spatial reliability maps while leaving the predicted class probabilities unchanged.

\subsection{Diffusion Segmentation and Reliability Assessment}

Diffusion models generate structured outputs through iterative denoising~\cite{ho2020denoising,austin2021structured,dieleman2022continuous,campbell2022continuous}.
For medical image segmentation, MedSegDiff predicts a label map conditioned on the input image, using dynamic conditional encoding across denoising steps and a frequency-domain parser to suppress high-frequency noise~\cite{wu2024medsegdiff}.
In categorical diffusion, each reverse step outputs a pixel-wise class distribution, making the evolution of the prediction observable before the terminal state.
The sequence of intermediate distributions records how class probabilities change as the noisy label state is refined.

Diffusion models have also been studied for ambiguous segmentation and reliability assessment.
MoDiff represents inter-annotator ambiguity with probability-based label maps and generates diverse segmentation samples while retaining anatomical boundaries~\cite{ahn2025modiff}.
Its learnable frequency filter extracts morphology-related high-frequency information, while morphology-based cross-attention transfers structural cues into the denoising process.
CalDiff considers lesion segmentation with multiple annotations and multiple predictions~\cite{wang2025caldiff}.
Its step-wise calibration module handles uncertainty at individual denoising stages, and its sequence-aware module accounts for error accumulation across successive states.
The calibrated uncertainty maps identify low-confidence regions and support reliability assessment of the sampled lesion predictions.
Both approaches derive reliability from repeated sampling of the reverse process, whereas CARD compares two models along a single shared trajectory.

\begin{figure*}[!t]
\centering
\includegraphics[width=0.99\textwidth]{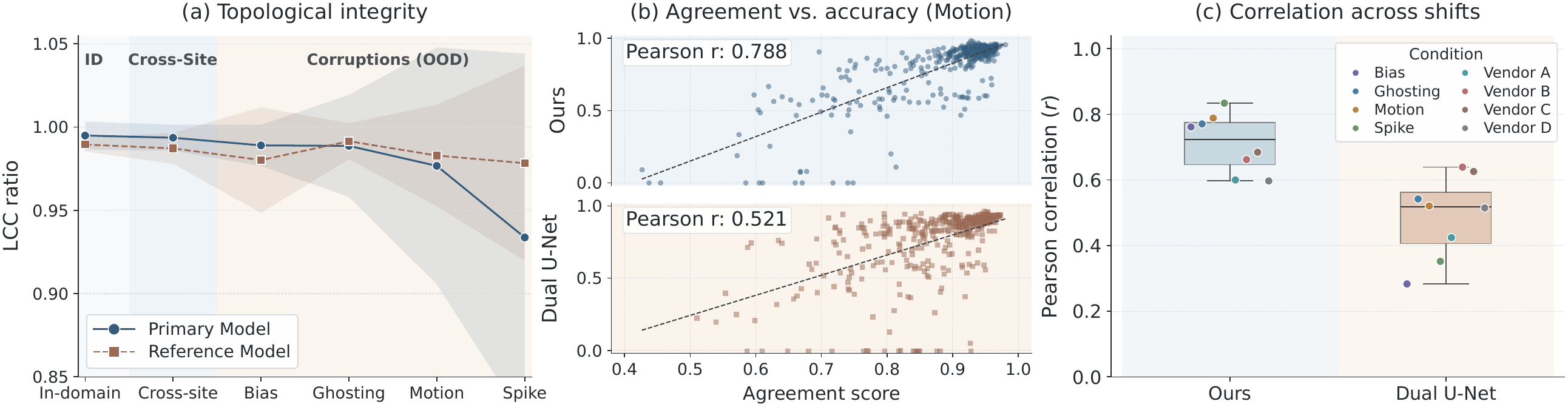}
\caption{\textbf{Disagreement tracks accuracy because only the primary fragments.}
\textbf{(a)} Largest connected component ratio. The primary loses connectivity as corruption worsens while the reference stays intact, so disagreement marks the primary's failures, not the reference's.
\textbf{(b)} Slice-level agreement against Dice under motion corruption, with two
matched discriminative U-Nets as a control.
\textbf{(c)} Pearson correlation across four artifact types and four vendor groups.}
\label{fig:agreement_validity}
\end{figure*}

\section{Motivation}
\label{sec:motivation}

\subsection{Preliminaries: Categorical Diffusion for Segmentation}
We represent the segmentation as a pixel-wise categorical distribution with $K$ classes.
At time step $t$, $\boldsymbol{x}_t$ denotes the categorical segmentation state over the image space $\Omega$, whose class-probability parameters form a tensor in $[0,1]^{H\times W\times K}$, with each pixel-wise vector summing to one.
Unlike Euclidean diffusion, we use categorical diffusion models for discrete data~\cite{austin2021structured}, with the class-probability vectors defined on the simplex~\cite{dieleman2022continuous,campbell2022continuous}.
At inference, the reverse process uses deterministic DDIM sampling~\cite{song2020denoising}.

\noindent \textbf{Forward Process.}
Let $\boldsymbol{x}_0$ denote the one-hot segmentation labels and
$\boldsymbol{u}=\boldsymbol{1}/K$ the uniform categorical distribution.
The forward process mixes $\boldsymbol{x}_0$ with $\boldsymbol{u}$ to define the distribution,
\begin{equation}
    \boldsymbol{x}_t \sim
    \mathrm{Cat}\!\left(
    \bar{\alpha}_t\boldsymbol{x}_0+
    (1-\bar{\alpha}_t)\boldsymbol{u}
    \right),
    \label{eq:forward_mix}
\end{equation}
where $\bar{\alpha}_t\in(0,1)$ is a predefined noise schedule, and $\mathrm{Cat}(\cdot)$ denotes a categorical distribution over $K$ classes.
As $\bar{\alpha}_t\rightarrow0$, the categorical distribution approaches
$\boldsymbol{u}$. At inference, we initialize the reverse process with the fixed uniform state $\boldsymbol{x}_T=\boldsymbol{u}$.

\noindent \textbf{Reverse Process.}
Starting from $\boldsymbol{x}_T^\theta=\boldsymbol{x}_T=\boldsymbol{u}$, A denoising network $f^\theta$ estimates the clean categorical distribution
from the current reverse state $\boldsymbol{x}_t^\theta$, conditioned on
the input MRI $\boldsymbol{I}$ and the step index,
\begin{equation}
    \hat{\boldsymbol{p}}^{\theta}_{t}
    =
    \mathrm{Softmax}\!\left(
    f^{\theta}(\boldsymbol{x}_t^\theta,\boldsymbol{I},t)
    \right).
\end{equation}
where $\hat{\boldsymbol{p}}^{\theta}_{t}(\boldsymbol{v})$ denotes the class
distribution at pixel $\boldsymbol{v}$.

Following the deterministic DDIM update~\cite{song2020denoising}, the
current state and predicted clean distribution are used to estimate the
residual and determine the next reverse state,
\begin{equation}
    \boldsymbol{x}_{t-1}^{\theta}
    =
    \arg\max_{k}
    \left[
    \bar{\alpha}_{t-1}\hat{\boldsymbol{p}}^{\theta}_{t}
    +(1-\bar{\alpha}_{t-1})
    \frac{
    \boldsymbol{x}_{t}^{\theta}
    -\bar{\alpha}_{t}\hat{\boldsymbol{p}}^{\theta}_{t}
    }{
    1-\bar{\alpha}_{t}
    }
    \right],
    \label{eq:reverse_ddim}
\end{equation}
where $k\in\{1,\ldots,K\}$ indexes the segmentation classes, and the
argmax is applied pixel-wise over the class dimension, with the resulting
labels represented as categorical states for the next reverse step.
With $\boldsymbol{x}_T$ fixed and Eq.~\eqref{eq:reverse_ddim} deterministic, the
trajectory $\{\boldsymbol{x}_t^\theta\}_{t=0}^{T}$ is determined entirely by the
input image, so two denoisers evaluated along it differ only through their
own predictions, free of sampling variance.
This sequence provides a temporal axis along which the reliability of
the intermediate states can be assessed.

\noindent \textbf{Temperature scaling.}
Given pixel-wise logits $\boldsymbol{z}(\boldsymbol{v})\in\mathbb{R}^{K}$, temperature scaling divides the logits by a temperature $\tau(\boldsymbol{v})>0$ before the softmax,
\begin{equation}
    \boldsymbol{q}(\boldsymbol{v})=\mathrm{Softmax}\!\left(\boldsymbol{z}(\boldsymbol{v})/\tau(\boldsymbol{v})\right),
    \label{eq:temperature_scaling}
\end{equation}
\label{temperature_scaling}
so that $\boldsymbol{q}(\boldsymbol{v})$ better reflects the likelihood that the prediction at that pixel is correct.
Standard temperature scaling uses a single global scalar fitted on a labeled validation set; spatial variants predict a temperature map.

\subsection{Agreement-on-the-Line in Categorical Diffusion}

This section substantiates the two design choices behind CARD: a capacity-limited reference model, and aggregation of disagreement along the reverse trajectory.
Both rest on the ability of generative models to represent anatomical shape explicitly.
A deliberately weakened model can provide a useful contrast: Karras et al.~\cite{karras2024guiding} show that guiding diffusion sampling with a smaller, undertrained copy of the same model improves sample quality, since the degraded copy retains coarse structure but lacks fine detail. We repurpose the same capacity asymmetry for calibration, using the weaker model as a static comparator.
We hypothesize that restricting the reference capacity preserves dominant anatomical structure while maintaining sufficient difference from the high-capacity primary to make disagreement informative, which we call prediction contrast.
The reference then acts as a structurally stable comparator: a high-capacity reference reproduces the primary's local failure modes, whereas an overly weak reference introduces disagreement through its own errors. 
Reliable calibration accordingly requires two properties: (i) topological stability of the reference prediction, and (ii) inter-model agreement scores that reliably track segmentation quality under domain shift.
Fig.~\ref{fig:agreement_validity} evaluates the first through foreground connectivity and the second through the agreement-Dice relationship across corruption and cross-vendor settings.

\noindent \textbf{Connectivity robustness through generative shape priors}
Domain shifts and artifacts can cause discriminative models to produce fragmented predictions.
We quantify foreground connectivity using the largest connected component (LCC) ratio~\cite{Giarratano2020OCTA}: for a predicted foreground $\hat{S}$, the fraction of the predicted area occupied by the single largest continuous structure, $|\hat{S}_{\mathrm{LCC}}|/|\hat{S}|$. 
As shown in Fig.~\ref{fig:agreement_validity}(a), the capacity-limited generative reference fragments less than the primary under severe shift. 
Under spike corruption the LCC ratio is 0.978 for the reference and 0.934 for the primary, showing that the reference retains the dominant foreground component even when local appearance is corrupted.
Where local evidence remains reliable the two predictions stay close, and larger discrepancies emerge around corruption-induced errors as the primary deviates from the reference.

Two aspects of generative training account for this behavior.
First, the denoiser is trained to reconstruct a complete label map from a noised state.
At large $t$ the state carries little label information, so the network increasingly relies on a learned prior over plausible segmentations to recover the missing structure.  
Restricting capacity restricts what that prior can encode, retaining dominant low-frequency anatomical modes while discarding fine boundary detail.
Because MRI artifacts perturb mainly high-frequency image content, a reference that represents only coarse structure is comparatively insensitive to them, which is the behavior measured in Fig.~\ref{fig:agreement_validity}(a).
Second, the primary and the reference occupy different points of the bias-variance trade-off.
The primary is evidence-dominated: accurate in domain, but its errors grow wherever local appearance is corrupted.
The reference is prior-dominated: less accurate overall, but its error is a property of the learned anatomical prior and therefore changes little with acquisition.
In domain the two agree because both are accurate; under shift the primary degrades locally while the reference does not, so their disagreement localizes the primary's evidence-driven failures.
A matched pair of discriminative models lacks this asymmetry.
Two U-Nets trained on the same data share architecture-induced failure modes, so their errors remain correlated under corruption and they continue to agree where both are wrong.

This account also predicts a limit.
As reference capacity falls further, its own bias eventually dominates the comparison, and disagreement increasingly captures errors introduced by the reference.
Sec.~\ref{subsec:reference_capacity} examines this trade-off empirically.

\noindent \textbf{Agreement as quality proxy}
We next evaluate whether the primary--reference disagreement can serve as a proxy for segmentation quality. 
Unlike discriminative segmentors, which provide only a terminal prediction, diffusion exposes pixel-wise class distributions throughout the entire denoising trajectory. 
The ordered sequence distinguishes disagreement confined to a single reverse step from disagreement that persists as the categorical state is refined. 
We aggregate the pixel-wise divergence across these states to form the persistent disagreement map $\bar{d}(\boldsymbol{v})$, defined in Sec.~\ref{subsec:tac}.
To compare this map with the Dice score, we average the normalized disagreement over the analysis region $\mathcal{R}\subseteq\Omega$ to obtain a scalar agreement score,
\begin{equation}
    \mathrm{Agr}
    =1-\frac{1}{|\mathcal{R}|\ln2}\sum_{\boldsymbol{v}\in\mathcal{R}}\bar{d}(\boldsymbol{v}).
    \label{eq:agreement_score}
\end{equation}
Since the Jensen--Shannon (JS) divergence with natural logarithms is bounded above by $\ln 2$, the score lies in $[0,1]$, with higher values indicating stronger agreement.
Eq.~\eqref{eq:agreement_score} is used for the analyses in Fig.~\ref{fig:agreement_validity}(b) and (c), while $\bar{d}(\boldsymbol{v})$ retains the spatial detail used for calibration.

Dual U-Net control consists of two discriminative U-Nets trained with the same source split, preprocessing, losses, and optimization schedule as the diffusion models.
We compute agreement between their terminal predictions to test whether dual-model agreement alone tracks segmentation quality. 
In the motion-corruption case study in Fig.~\ref{fig:agreement_validity}(b), trajectory agreement attains a Pearson correlation of 0.788 with the ground-truth Dice, against 0.521 for Dual U-Net control~\cite{benesty2009pearson}.
Fig.~\ref{fig:agreement_validity}(c) extends this comparison to four artifact types and four vendor groups, and trajectory agreement is higher in all eight settings. 
The margin there isolates the contribution of the ordered generative process beyond dual-model agreement.

Each reverse step re-imposes the same prior on the current state.
A discrepancy produced by transient ambiguity at one state is typically resolved at the next, whereas a discrepancy arising from a persistent conflict between image evidence and anatomical plausibility survives repeated refinement.
Aggregating over the trajectory separates the two, while the terminal state records only their sum.
A discriminative model exposes a single prediction and no intermediate refinement, so this separation is unavailable to the Dual U-Net control however its agreement is measured.
The next section therefore maps the aggregated disagreement to a spatial calibration strength, so that larger discrepancies attenuate confidence more strongly.


\section{Method}
\label{sec:method}

\begin{figure*}[!t]
\centering
\includegraphics[width=\textwidth]{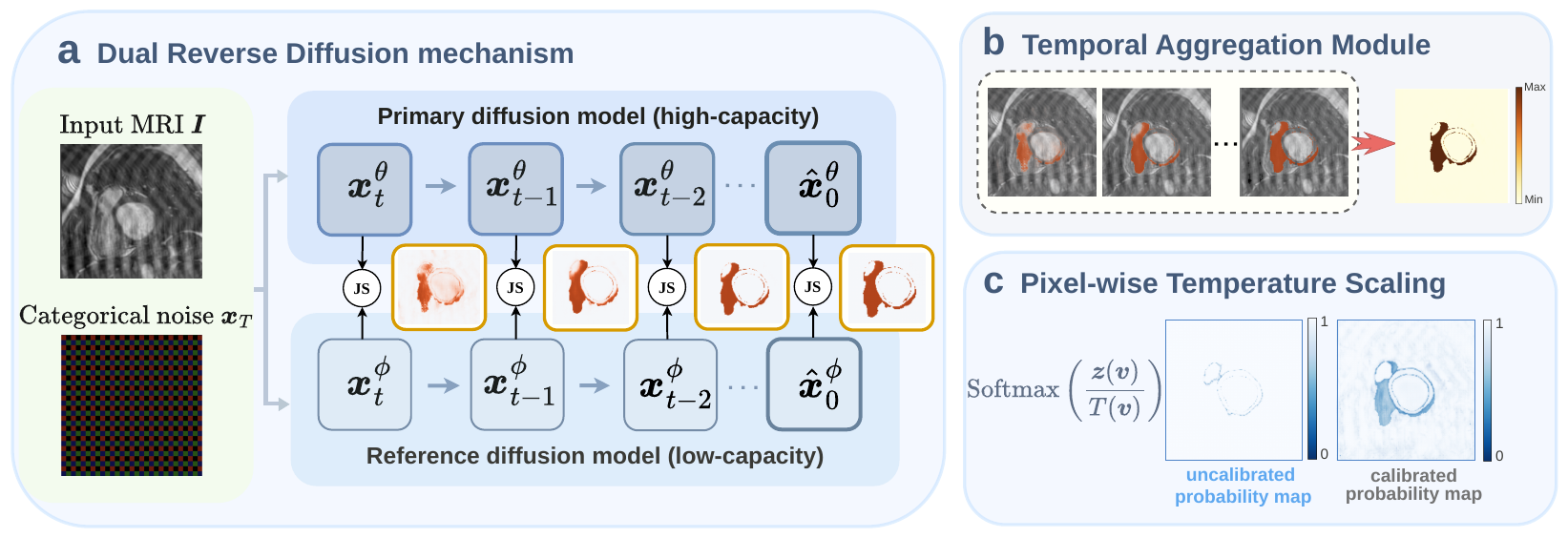}
\caption{\textbf{Overview of CARD.}
(a) Dual reverse diffusion.
From input MRI $\boldsymbol{I}$, the primary and reference denoisers follow paired DDIM-style reverse trajectories, initialized from the same categorical noise state $\boldsymbol{x}_T$ and evolving through intermediate states $\boldsymbol{x}_t^\theta$ and $\boldsymbol{x}_t^\phi$ to predicted clean states $\hat{\boldsymbol{x}}_0^\theta$ and $\hat{\boldsymbol{x}}_0^\phi$, respectively.
At each step $t$, these states define the pixel-wise categorical distributions $\hat{\boldsymbol{p}}^\theta_t(\boldsymbol{v})$ and $\hat{\boldsymbol{p}}^\phi_t(\boldsymbol{v})$ over $K$ classes.
Their pixel-wise disagreement is denoted $d_t(\boldsymbol{v})$ and computed by JS divergence.
(b) Temporal aggregation.
We aggregate ${d_t(\boldsymbol{v})}_{t\in\mathcal{T}}$ over the sampled reverse states to obtain a stable pixel-wise signal.
(c) Pixel-wise temperature scaling.
The aggregated disagreement defines a temperature map $\tau(\boldsymbol{v})$, which calibrates the final primary logits by $\mathrm{Softmax}(\boldsymbol{z}(\boldsymbol{v})/\tau(\boldsymbol{v}))$.}

\label{fig:overview}
\end{figure*}

\subsection{Reference-Guided Dual Diffusion}
\label{subsec:agreement}


We pair a high-capacity primary denoiser $f^\theta$ with a capacity-limited reference denoiser $f^\phi$, both conditioned on the same input image $\boldsymbol{I}$ and trained on the same source segmentation task.
At each sampled reverse step $t$, the two denoisers are evaluated on the same intermediate state $\boldsymbol{x}_t$ and output pixel-wise categorical distributions $\hat{\boldsymbol{p}}^\theta_t(\boldsymbol{v})$ and $\hat{\boldsymbol{p}}^\phi_t(\boldsymbol{v})$ over $K$ classes.
Only the primary determines the next sampled categorical state; the reference is evaluated on the shared trajectory but never advances it.
The reference capacity is constrained through a reduced architecture and a truncated training schedule, so that it encodes global anatomical structure rather than transient local detail (Sec.~\ref{sec:motivation}).

We measure the step-wise disagreement using JS divergence~\cite{lin1991divergence},
\begin{equation}
    d_t(\boldsymbol{v})
    =\mathrm{JS}\!\left(
    \hat{\boldsymbol{p}}^{\theta}_{t}(\boldsymbol{v}),\,
    \hat{\boldsymbol{p}}^{\phi}_{t}(\boldsymbol{v})
    \right)
    =\tfrac{1}{2}\mathrm{KL}(\hat{\boldsymbol{p}}^{\theta}_{t}\|\boldsymbol{m}_t)
    +\tfrac{1}{2}\mathrm{KL}(\hat{\boldsymbol{p}}^{\phi}_{t}\|\boldsymbol{m}_t),
    \label{eq:js}
\end{equation}
where $\boldsymbol{m}_t=(\hat{\boldsymbol{p}}^{\theta}_{t}+\hat{\boldsymbol{p}}^{\phi}_{t})/2$.
The divergence is symmetric and, with natural logarithms, bounded above by $\ln 2$ irrespective of $K$, so $d_t$ is comparable across reverse steps and across tasks.
Probabilities are clamped at $\epsilon_{\mathrm{JS}}=10^{-12}$ and renormalized before evaluation.
Because the reference is evaluated on states the primary has already produced, CARD requires one additional forward pass per sampled step and no additional sampling of the reverse process.

\subsection{Temporal Aggregation Calibration}
\label{subsec:tac}

Disagreement at a single reverse step may reflect step-specific noise or transient ambiguity in the current categorical state.
Temporal Aggregation Calibration (TAC) averages disagreement over the sampled reverse trajectory.
\begin{equation}
    \bar d(\boldsymbol{v})
    =
    \frac{1}{|\mathcal{T}|}
    \sum_{t\in\mathcal{T}} d_t(\boldsymbol{v}),
    \label{eq:tac_average}
\end{equation}
where $\mathcal{T}\subseteq\{0,\ldots,T\}$ is the set of sampled reverse steps used at inference.
Since $\bar{d}$ is a mean of quantities bounded by $\ln 2$, it inherits the same bound.

The aggregated disagreement is mapped to a bounded spatial temperature field:
\begin{equation}
    \temp(\boldsymbol{v})
    =
    \temp_{\min}
    +
    \sigma\!\left(
    \frac{\bar d(\boldsymbol{v})-w_b}{w_k}
    \right)
    (\temp_{\max}-\temp_{\min}),
    \label{eq:agreement_to_temp}
\end{equation}
where $\sigma(\cdot)$ is the logistic function, $1\leq\temp_{\min}\leq\temp_{\max}$, and $w_b,w_k>0$ control the location and width of the transition.
The mapping is monotonically increasing, so it preserves ordering of disagreement strength, and $\temp$ remains within $(\temp_{\min},\temp_{\max})$, which limits the magnitude of any single correction.

Let $\logits(\boldsymbol{v})\in\R^K$ denote the terminal primary logits, taken from the final reverse step.
One positive scalar temperature is applied to all $K$ logits at a pixel, so Eq.~\eqref{eq:temperature_scaling} preserves their ordering and leaves the predicted label unchanged; CARD alters confidence without altering the segmentation.
Constraining $\temp_{\min}\geq1$ restricts the correction to attenuation, so confidence can be reduced but never increased.
This is deliberate: the failure mode targeted here is overconfidence under shift, and an unconstrained field could raise confidence in regions where the primary is wrong.

\noindent\emph{Hyper-parameter selection.}
The mapping parameters $\{w_b,w_k,\tau_{\min},\tau_{\max}\}$ are fitted with both denoisers frozen, so no gradient reaches the segmentor.
Fitting uses a pool containing the original source-validation images together with three label-preserving intensity perturbations, namely contrast adjustment, gamma transformation, and additive Gaussian receiver noise,
following prior work that uses image transformations and acquisition noise to estimate prediction susceptibility and image-based uncertainty in medical image segmentation~\cite{ouyang2022improved,wang2019aleatoric}.
For each candidate setting, the aggregated disagreement is mapped to a temperature field and applied to the terminal primary logits, and the objective is the mean pixel-wise negative log likelihood (NLL) over the original and augmented validation images.
A coarse grid supplies the initialization for constrained SLSQP refinement~\cite{kraft1988slsqp}.
The pool is constructed from source-validation data only and excludes the artifact corruptions and acquisition shifts used for held-out testing.

\section{Experimental Protocol}
\label{sec:experimental_protocol}

\subsection{Datasets}
\label{sec:applications_data}

We evaluate on three tasks, each with a source cohort and one or more held-out shifted cohorts.
Only the ACDC-C corruptions are synthetic; the remaining shifts arise from real differences in scanner, vendor, or acquisition site.

\noindent\textbf{Cardiac.}
Cardiac segmentation distinguishes the right ventricle, myocardium, left ventricle, and background in cine MRI. 
ACDC~\cite{bernard2018acdc} is the source cohort, using the patient-level split of prior work~\cite{Bai_2023_CVPR}, with 70 subjects for training, 10 for source validation, and 20 for source testing.
ACDC-C provides a controlled artifact-shift by applying bias-field, motion, ghosting, and k-space spike corruptions to the 160 ACDC test images with TorchIO~\cite{perezgarcia2021torchio}, which preserves the original anatomy and labels and is applied only at evaluation. 
M\&Ms~\cite{campello2021mnms} contains 690 multi-center, multi-vendor cine bSSFP scans and is reserved for real cross-vendor evaluation of models trained on ACDC.

\noindent\textbf{Prostate.}
Prostate segmentation separates the gland from the background in axial T2-weighted MRI. 
PROSTATEx with Meyer/Seg-HiRes annotations is the source cohort~\cite{litjens2017prostatex,meyer2018automatic}, with 16 cases for training, 4 for source validation, and 20 for source testing.
The 50 annotated PROMISE12 training cases~\cite{litjens2014promise12}, acquired across institutions with different scanners and acquisition parameters, form the held-out cohort.

\noindent\textbf{Brain lesion.}
Brain lesion segmentation targets sparse stroke lesions in ATLAS, a multi-site T1-weighted post-stroke MRI dataset with manual lesion masks~\cite{liew2022atlas}. 
Scanner metadata defines 13 strata in the processed labeled cohort.
SiemensTrioTim is the source stratum, with 59 cases for training, 13 for source validation, and 16 for source testing, while the remaining 565 cases from 12 held-out strata are used for scanner-shift evaluation, grouped by manufacturer as Philips, Siemens, and GE.

\noindent\textbf{Preprocessing.}
The in-plane matrix is center cropped or padded to $256 \times 256$ pixels for cardiac and prostate images and $224 \times 224$ pixels for brain. 
PROSTATEx and PROMISE12 are resampled to $0.8 \times 0.8 \times 1.5$~mm, and all tasks use per volume z-score normalization.

\subsection{Baselines}

Segmentation models are trained on source-domain data and calibration parameters are fitted on the corresponding source-validation splits. Held-out cohorts are used for all evaluations reported in Sec.~\ref{sec:results}.

Conventional baselines are applied to deterministic SwinUNETR segmentors ~\cite{hatamizadeh2021swinunetr}, while CARD operates on a time-conditioned SwinUNETR diffusion segmentor and uses its reverse states for calibration. 
UC is the frozen deterministic model without calibration. 
Alea. is trained separately following aleatoric uncertainty modeling~\cite{kendall2017uncertainties, wang2019aleatoric}: it predicts class-wise log variance and averages predictions sampled from the resulting logit distribution.

Following the grouping in Sec.~\ref{sec:related}, the static baselines are TS and ETS~\cite{guo2017calibration,zhang2020mix}, the output-conditioned baselines are LTS~\cite{ding2021local}, HTS~\cite{balanya2024adaptive}, and PTS~\cite{tomani2022parameterized}, and CalibNet~\cite{ouyang2022improved} is the image-conditioned baseline. 
We additionally apply ETS and HTS to the terminal logits of the same primary diffusion segmentor used by CARD, denoted as Primary+ETS and Primary+HTS. 
These backbone-matched controls separate the contribution of the diffusion backbone from that of trajectory disagreement.
Baselines requiring validation-based selection use the same augmented source-validation pool as CARD.

\begin{table*}[!t]
\begin{center}
\centering
\refstepcounter{table}\label{tab:main_calibration}
{\footnotesize TABLE~\thetable\\
\textbf{Calibration results across cardiac, prostate, and brain MRI shift settings.}
Methods are grouped by what the calibrator conditions on at deployment. Cells report ECE/SCE/ACE in percentage points; NLL is averaged with equal weight across the conditions in each block. All metrics are computed within the ROI; lower is better. Best per metric component in \textbf{bold}, second best \underline{underlined}.\par}
\vspace{3pt}
\scriptsize
\setlength{\tabcolsep}{3pt}
\renewcommand{\arraystretch}{1.05}
\resizebox{\textwidth}{!}{%
\begin{tabular}{@{}l|ccccc|>{\centering\arraybackslash}m{0.65cm}|cccc|>{\centering\arraybackslash}m{0.65cm}}
\toprule
& \multicolumn{11}{c}{\textbf{(a) Cardiac artifact and cross-vendor shifts}}\\[0.3ex]
\cmidrule(lr){2-12}
& \multicolumn{5}{c}{ACDC-C artifact}
&
& \multicolumn{4}{c}{M\&Ms vendor}
& \\
\textbf{Method}
& Clean (ID) & Bias & Mot. & Ghost & Spike
& NLL$\downarrow$
& A (Sie.) & B (Phi.) & C (GE) & D (Can.)
& NLL$\downarrow$ \\
\midrule
\gA{No post-hoc calibration}
\M{UC}
& 6.36/3.28/2.71
& 11.40/5.89/5.34
& 9.75/5.00/4.29
& 11.40/5.86/5.24
& 24.00/12.31/11.75
& 0.945
& 16.79/8.61/7.87
& 10.16/5.20/4.45
& 8.48/4.37/3.67
& 11.09/5.74/5.04
& 0.792 \\

\M{Alea.~\cite{kendall2017uncertainties}}
& 5.96/3.06/2.55
& 10.95/5.66/5.15
& 10.10/5.20/4.54
& 10.49/5.41/4.80
& 23.64/12.05/11.50
& 1.037
& 16.69/8.52/7.82
& 10.01/5.13/4.47
& 8.23/4.22/3.61
& 10.52/5.39/4.72
& 0.864 \\

\gA{Static}
\M{TS~\cite{guo2017calibration}}
& 4.57/2.57/2.32
& 9.11/5.07/4.86
& 7.71/4.14/3.74
& 9.32/5.01/4.71
& 21.87/11.52/11.25
& 0.645
& 14.37/7.66/7.26
& 7.98/4.23/3.80
& 6.18/3.38/3.03
& 8.83/4.80/4.42
& 0.548 \\

\M{ETS~\cite{zhang2020mix}}
& 2.70/2.24/2.22
& 6.69/4.66/4.65
& 5.36/3.42/3.32
& 7.09/4.45/4.35
& 19.57/10.91/10.81
& \underline{0.539}
& 11.61/6.99/6.83
& 5.37/3.37/3.23
& 3.55/2.78/2.72
& 6.23/4.02/3.94
& \underline{0.469} \\

\gA{Output-conditioned}
\M{HTS~\cite{balanya2024adaptive}}
& 2.58/\underline{2.21}/2.14
& 6.42/4.63/4.57
& 5.15/3.38/3.27
& 6.73/4.42/4.30
& 19.24/10.86/10.77
& 0.623
& 11.24/6.93/6.82
& \textbf{5.16}/3.36/3.21
& 3.69/\underline{2.75}/2.69
& \underline{5.93}/4.00/3.91
& 0.531 \\

\M{PTS~\cite{tomani2022parameterized}}
& \underline{2.37}/\underline{2.21}/2.24
& 6.69/4.65/4.65
& 4.90/3.32/3.28
& 6.65/4.36/4.30
& 19.50/10.96/10.88
& 0.542
& 11.71/6.95/6.79
& 5.30/\underline{3.34}/3.25
& \underline{3.54}/\textbf{2.71}/\underline{2.69}
& 6.20/3.99/3.96
& 0.471 \\

\M{LTS~\cite{ding2021local}}
& 2.56/\underline{2.21}/2.16
& 5.55/4.39/4.39
& \textbf{3.76}/\underline{3.09}/\underline{3.03}
& 6.52/3.99/3.95
& 18.48/10.70/10.66
& 0.600
& 11.75/6.95/6.78
& \underline{5.23}/3.35/3.22
& 3.51/2.76/2.69
& 6.08/\underline{3.94}/3.88
& 0.522 \\

\gA{Image-conditioned}
\M{CalibNet~\cite{ouyang2022improved}}
& \textbf{2.27}/2.31/2.22
& \underline{5.21}/4.65/4.58
& \underline{4.58}/3.30/3.17
& \underline{6.44}/\underline{3.83}/\underline{3.62}
& \underline{16.22}/10.20/10.10
& 0.570
& \underline{10.62}/\underline{6.65}/6.47
& 5.60/3.51/3.32
& 3.61/2.79/2.72
& 6.39/4.08/3.93
& 0.523 \\

\gA{Backbone-matched control (calibrators on primary diffusion logits)}
\M{Primary+ETS~\cite{zhang2020mix}}
& 5.87/2.97/2.32
& 7.82/3.97/3.33
& 8.46/4.25/3.47
& 9.73/4.90/4.12
& 18.43/\underline{9.29}/\underline{8.70}
& 0.559
& 14.59/7.33/\underline{6.47}
& 9.76/4.90/3.93
& 7.54/3.82/3.01
& 9.42/4.73/\underline{3.84}
& 0.562 \\

\M{Primary+HTS~\cite{balanya2024adaptive}}
& 5.87/3.00/2.46
& 8.72/4.36/3.49
& 9.15/4.58/3.62
& 10.45/5.23/4.27
& 19.40/9.71/8.95
& 0.647
& 14.51/7.36/6.78
& 9.63/4.87/4.24
& 7.43/3.79/3.29
& 9.26/4.70/4.08
& 0.635 \\

\gA{Trajectory-conditioned}
\M{\textbf{CARD (ours)}}
& 2.81/\textbf{2.08}/\textbf{1.86}
& \textbf{4.00}/\textbf{3.03}/\textbf{2.91}
& 4.91/\textbf{3.00}/\textbf{2.73}
& \textbf{6.20}/\textbf{3.62}/\textbf{3.32}
& \textbf{13.81}/\textbf{7.61}/\textbf{7.42}
& \textbf{0.405}
& \textbf{9.86}/\textbf{5.68}/\textbf{5.42}
& 5.41/\textbf{3.27}/\textbf{2.92}
& \textbf{3.45}/2.77/\textbf{2.52}
& \textbf{5.05}/\textbf{3.41}/\textbf{3.13}
& \textbf{0.437} \\

\midrule
\midrule
& \multicolumn{6}{c|}{\textbf{(b) Prostate source-to-target transfer}}
& \multicolumn{5}{c}{\textbf{(c) Brain scanner shift}} \\
\cmidrule(lr){2-7}\cmidrule(lr){8-12}
\textbf{Method} & \multicolumn{2}{c}{PROSTATEx (ID)}
& \multicolumn{3}{c|}{PROMISE12$^\dagger$}
& NLL$\downarrow$
& SiemensTrioTim (ID)
& Philips$^\ddagger$
& Siemens$^\ddagger$
& GE$^\ddagger$
& NLL$\downarrow$ \\
\midrule

\gA{No post-hoc calibration}
\M{UC}
& \multicolumn{2}{c}{18.04/18.54/18.00}
& \multicolumn{3}{c|}{28.95/29.68/29.07}
& 1.786
& 13.61/13.87/13.10
& 22.24/22.40/21.59
& 18.82/19.02/18.48
& 13.55/13.72/13.06
& 1.587 \\

\M{Alea.~\cite{kendall2017uncertainties}}
& \multicolumn{2}{c}{13.36/\underline{13.82}/\underline{13.18}}
& \multicolumn{3}{c|}{24.70/\underline{25.29}/\underline{24.73}}
& 1.553
& 12.01/12.42/12.01
& 20.31/20.59/\underline{20.11}
& 17.39/17.73/17.31
& 11.57/\underline{11.91}/\underline{11.48}
& 1.223 \\

\gA{Static}
\M{TS~\cite{guo2017calibration}}
& \multicolumn{2}{c}{14.69/16.20/16.20}
& \multicolumn{3}{c|}{24.92/26.93/26.80}
& 0.947
& 11.42/12.06/11.77
& 21.20/21.52/21.22
& 17.51/18.02/17.84
& 12.60/12.94/12.66
& 0.941 \\

\M{ETS~\cite{zhang2020mix}}
& \multicolumn{2}{c}{12.71/15.49/15.64}
& \multicolumn{3}{c|}{22.72/25.82/25.81}
& \underline{0.808}
& 9.24/10.71/10.58
& 20.05/\underline{20.56}/20.35
& 16.13/\underline{17.09}/\underline{16.97}
& 11.47/12.11/11.96
& \underline{0.732} \\

\gA{Output-conditioned}
\M{HTS~\cite{balanya2024adaptive}}
& \multicolumn{2}{c}{\underline{10.38}/15.62/15.66}
& \multicolumn{3}{c|}{\underline{20.05}/25.43/25.44}
& 0.880
& \underline{8.55}/\underline{10.56}/\underline{10.50}
& 19.92/20.65/20.49
& \underline{15.92}/17.20/17.10
& 11.45/12.29/12.13
& 0.996 \\

\M{PTS~\cite{tomani2022parameterized}}
& \multicolumn{2}{c}{10.93/15.51/15.58}
& \multicolumn{3}{c|}{20.53/25.42/25.43}
& 0.829
& 10.58/11.49/11.27
& 20.76/21.14/20.88
& 16.97/17.65/17.50
& 12.17/12.60/12.38
& 0.829 \\

\M{LTS~\cite{ding2021local}}
& \multicolumn{2}{c}{11.95/17.23/17.26}
& \multicolumn{3}{c|}{22.65/26.78/26.74}
& 1.043
& 8.63/10.81/10.78
& \underline{19.84}/20.90/20.77
& 16.18/17.49/17.41
& \underline{11.24}/12.28/12.20
& 0.989 \\

\gA{Image-conditioned}
\M{CalibNet~\cite{ouyang2022improved}}
& \multicolumn{2}{c}{14.05/17.00/16.96}
& \multicolumn{3}{c|}{23.98/26.82/26.73}
& 0.974
& 10.48/11.68/11.89
& 21.20/21.34/21.33
& 16.02/16.19/16.22
& 11.67/12.71/12.88
& 0.873 \\

\gA{Backbone-matched control (calibrators on primary diffusion logits)}
\M{Primary+ETS~\cite{zhang2020mix}}
& \multicolumn{2}{c}{16.53/16.55/15.35}
& \multicolumn{3}{c|}{33.31/33.36/32.46}
& 1.199
& 13.68/13.69/12.79
& 23.34/23.34/22.60
& 20.24/20.31/19.73
& 14.49/14.50/13.87
& 0.974 \\

\M{Primary+HTS~\cite{balanya2024adaptive}}
& \multicolumn{2}{c}{16.06/16.23/15.51}
& \multicolumn{3}{c|}{33.24/33.48/32.78}
& 1.359
& 13.76/13.78/13.22
& 23.59/23.61/22.92
& 20.39/20.49/20.00
& 14.87/14.87/14.24
& 1.185 \\

\gA{Trajectory-conditioned}
\M{\textbf{CARD (ours)}}
& \multicolumn{2}{c}{\textbf{6.42}/\textbf{11.18}/\textbf{11.29}}
& \multicolumn{3}{c|}{\textbf{19.97}/\textbf{23.43}/\textbf{23.41}}
& \textbf{0.565}
& \textbf{5.91}/\textbf{8.50}/\textbf{8.58}
& \textbf{15.91}/\textbf{16.93}/\textbf{16.85}
& \textbf{13.00}/\textbf{14.63}/\textbf{14.66}
& \textbf{7.70}/\textbf{8.93}/\textbf{9.25}
& \textbf{0.548} \\

\bottomrule
\end{tabular}
}
\begin{flushleft}
\footnotesize
\emph{Backbone-matched controls} calibrate the primary segmentor's terminal logits, separating trajectory agreement from the backbone.
ID denotes source-domain evaluation. 
$^\dagger$held-out PROMISE12. $^\ddagger$scanner-shift ATLAS.
\end{flushleft}
\end{center}

\vspace{-2pt}
\end{table*}

\begin{figure*}[!t]
\centering
\includegraphics[width=0.95\textwidth,trim=0 4 0 4,clip]{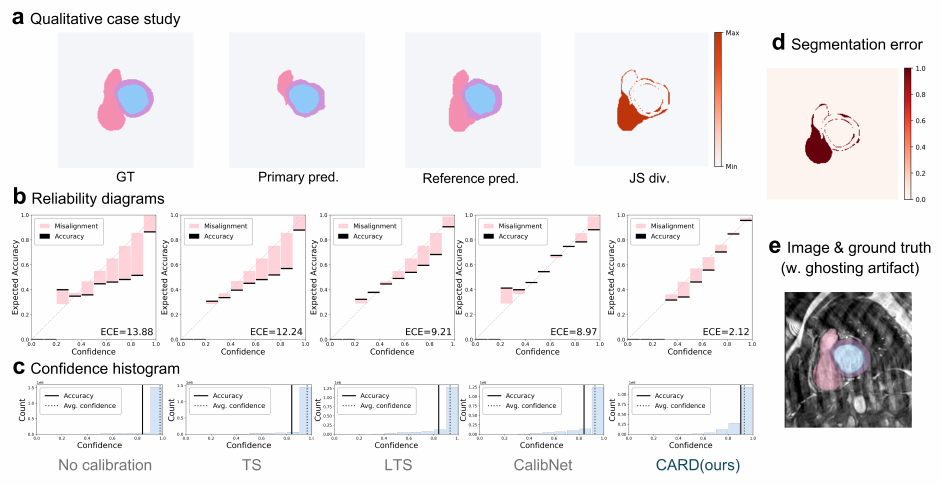}
\caption{\textbf{Trajectory disagreement lands where the primary is wrong, and calibrating on it closes the confidence gap.}
\textbf{(a)} Ground truth, primary and reference predictions, and the temporally aggregated JS disagreement for a ghosting-corrupted cardiac case. The disagreement map coincides with the primary's segmentation error in \textbf{(d)}, without access to labels.
\textbf{(b)} Reliability diagrams for the five calibration methods; shaded bars give the confidence-accuracy gap.
\textbf{(c)} Confidence histograms of the corresponding predictions.
\textbf{(d)} Primary segmentation error against the ground truth.
\textbf{(e)} Ghosting-corrupted input image with ground-truth contour.}
\label{fig:qualitative}
\end{figure*}

\subsection{Calibration Metrics}
\label{subsec:calibration_metrics}

To account for foreground--background imbalance, all metrics are evaluated within a region of interest $\mathcal{R}$ obtained by dilating the ground-truth foreground mask with a 10-pixel kernel~\cite{mehrtash2020confidence,ouyang2022improved}. 
We report Expected calibration error (ECE), Static calibration error (SCE), Adaptive calibration error (ACE), and NLL within $\mathcal{R}$, using $R=M=10$ bins.
NLL serves as both the fitting objective and an evaluation metric, so ECE, SCE, and ACE additionally provide criteria that no method optimizes directly.
Let $g(\boldsymbol{v})$ denote the ground-truth class and $q_k(\boldsymbol{v})$ the calibrated probability of class $k$.

ECE measures top-label calibration.
With confidence $c(\boldsymbol{v})=\max_k q_k(\boldsymbol{v})$ and correctness $a(\boldsymbol{v})=\mathds{1}[\arg\max_k q_k(\boldsymbol{v})=g(\boldsymbol{v})]$, pixels are partitioned into confidence bins $\{B_m\}_{m=1}^{M}$ and
\begin{equation}
    \ECE=
    \sum_{m=1}^M
    \frac{|B_m|}{|\OmegaSet|}
    \left|
    \operatorname{acc}(B_m)-\operatorname{conf}(B_m)
    \right|,
    \label{eq:ece}
\end{equation}
where $\operatorname{acc}$ and $\operatorname{conf}$ average $a$ and $c$ within a bin.

SCE extends the comparison to every class probability~\cite{nixon2019measuring}. 
With $B_{m,k}=\{\boldsymbol{v}\in\mathcal{R}:q_k(\boldsymbol{v})\in I_m\}$ for the $m$-th fixed interval $I_m$,
\begin{equation}
    \SCE=
    \frac{1}{K}
    \sum_{k=1}^{K}
    \sum_{m=1}^{M}
    \frac{|B_{m,k}|}{|\OmegaSet|}
    \left|
    \operatorname{acc}(B_{m,k})-
    \operatorname{conf}(B_{m,k})
    \right|.
    \label{eq:sce}
\end{equation}
where $\operatorname{acc}(B_{m,k})$ averages $\mathds{1}[g(\boldsymbol{v})=k]$ and $\operatorname{conf}(B_{m,k})$ averages $q_k(\boldsymbol{v})$.

ACE applies the same class-wise comparison to adaptive bins $A_{r,k}$ of approximately equal occupancy~\cite{nixon2019measuring}. 
\begin{equation}
    \ACE=
    \frac{1}{KR}
    \sum_{k=1}^{K}
    \sum_{r=1}^{R}
    \left|
    \operatorname{acc}(A_{r,k})-
    \operatorname{conf}(A_{r,k})
    \right|.
    \label{eq:ace}
\end{equation}


\subsection{Implementation Details}

The backbone is a time-conditioned SwinUNETR~\cite{hatamizadeh2021swinunetr}.
For each task the primary segmentor uses the Base configuration, with the checkpoint selected by source-validation Dice.
The Mini reference at 25k steps was fixed a priori and used unchanged across all three applications; the capacity and checkpoint analysis in
Sec.~\ref{subsec:reference_capacity} was performed retrospectively to assess the sensitivity of this choice.

Primary segmentors are optimized with Adam~\cite{kingma2014adam} at batch size 8 after a 1,000-step learning-rate warm-up.
We use a cosine schedule over a 50-step categorical diffusion horizon, and at inference a five-step deterministic reverse sampler evaluated at indices $[49,37,24,12,0]$, so $|\mathcal{T}|=5$ in Eq.~\eqref{eq:tac_average}. In implementation, we replace the estimated noise term in Eq.~\eqref{eq:reverse_ddim} with the uniform prior $\boldsymbol{u}$, which provides a sufficiently stable approximation in practice. The resulting state is then discretized by pixel-wise argmax.
The training objective combines a variational diffusion loss with an auxiliary clean-state prediction loss comprising cross-entropy, Dice, and boundary terms, and image condition is randomly masked with probability $0.2$.
Reference models used in the capacity analysis span 1.74M to 58.06M parameters and follow the same training and sampling protocol.

\section{Results and Discussion}
\label{sec:results}

\subsection{Calibration Results Across Tasks}

Table~\ref{tab:main_calibration} reports calibration under controlled cardiac artifacts, real cross-vendor cardiac shift, prostate source-to-target transfer, and brain scanner shift.
Among conventional calibration baselines, ETS gives the lowest mean NLL (0.539 ACDC-C, 0.469 M\&Ms, 0.808 prostate, 0.732 brain); PTS and HTS remain competitive on cardiac settings but do not match ETS elsewhere.
CARD lowers mean NLL further to 0.405, 0.437, 0.565, and 0.548 across the same four blocks. Applying ETS and HTS to the same primary diffusion segmentor used by CARD isolates the effect of the trajectory signal itself: relative to Primary+ETS, mean NLL falls from 0.559 to 0.405 (ACDC-C), 0.562 to 0.437 (M\&Ms), 1.199 to 0.565 (prostate), and 0.974 to 0.548 (brain). CARD also lowers ECE under spike corruption (16.22\% to 13.81\% versus CalibNet) and gives lower ECE, SCE, and ACE across prostate and brain groups.


This pattern indicates that calibration robustness under shift depends on whether the deployment-time signal stays aligned with the current error pattern. Static ETS is stable because its low-dimensional correction does not depend on case-specific cues. Output- and image-conditioned methods (HTS, PTS, CalibNet) are more case-specific: CalibNet is strongest under controlled cardiac artifacts, where perturbation susceptibility and its learned shape prior mirror the induced error, but its advantage shrinks under prostate and brain shift, where acquisition and morphology vary more. CARD instead measures disagreement between two diffusion models as the prediction forms, and this signal remains informative across all evaluated shifts. With the primary model fixed, it still beats ETS and HTS applied to the same terminal logits, attributing the gain to trajectory disagreement.

\subsection{Reliability Diagnostics Under Artifact Shift}

Fig.~\ref{fig:qualitative} tests whether trajectory disagreement localizes high-confidence errors under ghosting corruption.
The primary model produces a confident, localized segmentation error; the reference preserves more of the dominant foreground structure in that region.
The strongest JS responses occur exactly where the primary and reference predictions diverge, overlapping the ground-truth error in Fig.~\ref{fig:qualitative}(d), and the resulting temperature correction reduces confidence there.
For this case, global TS gives only a modest ECE reduction (13.88\% to 12.24\%); spatial calibrators LTS and CalibNet do better (9.21\%, 8.97\%), 
whereas CARD shifts mass toward lower-confidence bins, better matching observed accuracy.

This pixel-level result mirrors the aggregate calibration metrics: disagreement between the two diffusion paths tracks segmentation quality, not just at the case level but spatially, wherever local image evidence produces divergent predictions.

\begin{figure*}[!t]
\centering
\includegraphics[width=0.98\textwidth]{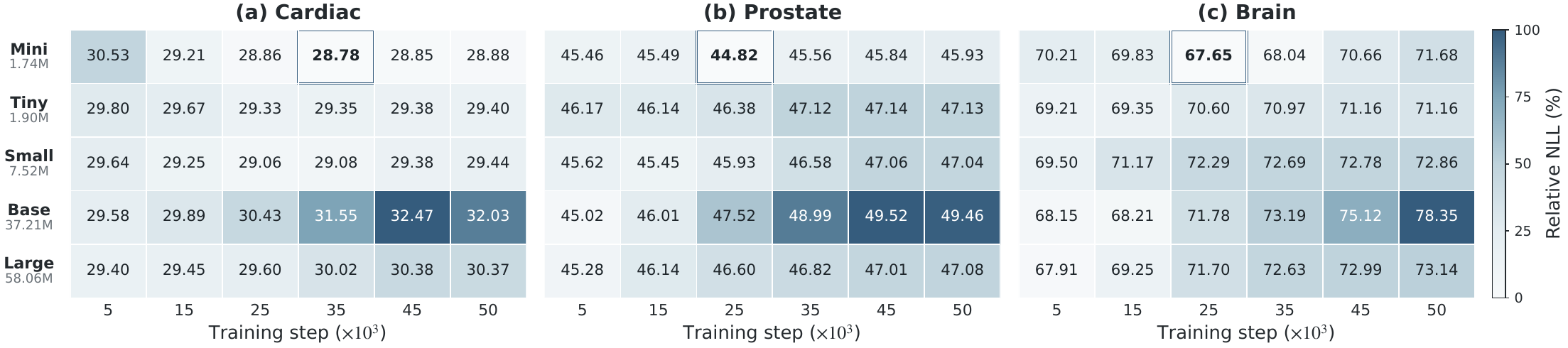}
\vspace{-2pt}
\caption{\textbf{Calibration favours a small, undertrained reference: the best NLL in every task comes from the smallest model, well before convergence.}
Panels (a) to (c): cardiac, prostate, and brain. Rows are reference models with parameter counts, columns are training steps, and cells report $\mathrm{NLL}\times100$ (lower is better, with the best cell per task outlined). The Base reference degrades steadily with further training, consistent with a high-capacity reference reproducing the primary’s failures and reducing prediction contrast. Colors are normalized within each panel.}
\label{fig:d1_nll_sensitivity}
\vspace{-4pt}
\end{figure*}

\subsection{Reference Capacity and Training Maturity}
\label{subsec:reference_capacity}

Fig.~\ref{fig:d1_nll_sensitivity} varies reference capacity (parameter count) and checkpoint maturity.
For cardiac and brain, the lowest NLL concentrates among Mini, Tiny, and Small references at early or intermediate checkpoints, while later Base and Large checkpoints generally worsen; prostate shows a broader low-NLL region.
Overall, increasing reference capacity or training does not consistently improve calibration, and the Mini reference used in the main experiments (adding 1.74M parameters) remains competitive across all tasks.

This suggests calibration benefits from a specific contrast regime, not reference strength alone: a reference too close to the primary reproduces its errors and collapses disagreement, while a very weak reference introduces disagreement from its own failures. The favorable region, where anatomy is preserved and predictions remain sufficiently distinct, explains why compact, undertrained references perform best, and reframes reference selection around useful error contrast instead of reference-model accuracy.
This mirrors the finding in diffusion guidance that a weaker copy of a model is more useful as a contrastive signal than as a standalone predictor~\cite{karras2024guiding}, though here the contrast is used to attenuate confidence without redirecting the sampling path.

\subsection{Computational Efficiency}
\label{subsec:computational_effi}
The primary denoiser contains 37.21M parameters; the Mini reference adds 1.74M parameters (a 4.68\% increase overall). At batch size 32, per-batch latency rises from 1736.8 to 2015.5\,ms on ACDC, 1734.5 to 2001.5\,ms on prostate, and 1219.7 to 1417.3\,ms on ATLAS (increases of 16.0\%, 15.4\%, and 16.2\%), with throughputs of 15.88, 15.99, and 22.58 slices/s. The overhead is modest relative to gains above.

\subsection{Component Ablation}

Table~\ref{tab:ablation} isolates which components of trajectory-conditioned calibration drive performance on ACDC-C under spike corruption.
Logit Mix (averaging primary and reference logits at every step) and Dual U-Net (the discriminative control from Sec.~\ref{sec:motivation}) test whether the calibration gain reduces to direct model fusion or terminal agreement between paired models; both underperform the full model (ECE 19.62\% and 19.22\% versus 13.81\%).
\textit{w/o} Ref. Contrast (identical denoiser on both paths) removes model contrast and raises ECE to 17.78\%.
Reducing temporal aggregation, tested with TAC-last (terminal state only) and TAC-mid (states $t\in\{24,12,0\}$), costs 3.81 and 3.64 points of ECE relative to using all five reverse steps, with the same ordering for SCE and ACE.

Together, these controls show the gain is specifically attributable to genuine primary-reference model contrast and to aggregating disagreement over the full reverse trajectory, not to logit fusion or paired-model agreement alone.

\begin{center}
\begin{minipage}{0.98\linewidth}
\centering
\refstepcounter{table}\label{tab:ablation}
{\footnotesize TABLE~\thetable\\
\textbf{Component ablation on ACDC-C spike corruption.}\par}
\vspace{3pt}
\scriptsize
\setlength{\tabcolsep}{6pt}
\begin{tabular}{lccc}
\toprule
Variant & ECE [\%] $\downarrow$ & SCE [\%] $\downarrow$ & ACE [\%] $\downarrow$ \\
\midrule
Primary only & 20.19 & 10.10 & 9.48 \\
Logit Mix & 19.62 & 9.82 & 9.03 \\
Dual U-Net & 19.22 & 11.49 & 11.10 \\
\textit{w/o} Ref. Contrast & 17.78 & 9.02 & 8.51 \\
\midrule
TAC-last & 17.62 & 8.96 & 8.47 \\
TAC-mid & 17.45 & 8.89 & 8.42 \\
\textbf{Proposed} & \textbf{13.81} & \textbf{7.61} & \textbf{7.42} \\
\bottomrule
\end{tabular}\par
\vspace{2pt}
{\footnotesize\raggedright
Upper block varies comparator: Primary only is the uncalibrated segmentor, Logit Mix averages primary and reference logits per step, Dual U-Net uses two matched discriminative models, and \textit{w/o Ref. Contrast} runs the primary on both paths.
Lower block varies aggregation: TAC-last uses $t=0$, TAC-mid averages over $t\in\{24,12,0\}$, and Proposed uses full trajectory.}
\end{minipage}
\end{center}

\subsection{Limitations and Future Work}

\label{sec:discussion}

CARD's central assumption, that primary and reference errors are largely independent, defines its main failure mode: shared errors can produce low disagreement in unreliable regions, thereby masking miscalibration instead of exposing it. The current temperature mapping only attenuates confidence, and evaluating the reference along the sampled trajectory adds inference cost (Sec.~\ref{subsec:computational_effi}). Future work includes reference selection beyond capacity, adaptive reverse-step sampling, correcting local underconfidence, and extending to 3D models.

\section{Conclusion}
\label{sec:conclusion}

We introduced trajectory agreement, a new test-time cue for calibrating diffusion-based medical image segmentation under domain shift. Rather than relying only on the terminal prediction, our method, CARD, exploits the fact that diffusion models expose an entire reverse trajectory of intermediate states, and shows that persistent disagreement between a primary model and a structurally stable, capacity-limited reference tracks segmentation reliability across artifact, vendor, and site shifts in cardiac, prostate, and brain MRI, where existing calibration methods are static or limited to a single output.

More broadly, this work suggests that reliability of a prediction is not fully captured by its final output alone: the process by which an iterative generative model arrives at that output carries additional evidence about when to trust it. Framing calibration around the generative trajectory offers a way to obtain reliability estimates from models that already produce a sequence of intermediate states, without architectural change. As diffusion and other iterative generative models become more common in high-stakes prediction settings, treating their generative dynamics as a first-class source of uncertainty may prove useful well beyond segmentation.

\ifdraftnotes
\fi

\bibliographystyle{IEEEtran}
\bibliography{references}

\end{document}